\documentclass[letterpaper, 10 pt, conference]{ieeeconf}  

\IEEEoverridecommandlockouts                              

\usepackage[space, compress, sort]{cite}

\usepackage{amsmath, amssymb, amsfonts, mathtools, amsthm}
\usepackage{algorithmic}
\usepackage{graphicx}
\usepackage{textcomp}
\usepackage{xcolor}
\usepackage{pifont}
\usepackage[linesnumbered,ruled]{algorithm2e}
\usepackage{paralist}
\usepackage[caption=false]{subfig}
\usepackage{bm}
\usepackage{scalerel}
\usepackage{multirow}
\usepackage{array, makecell} %
\theoremstyle{plain}
   
\theoremstyle{remark}

\theoremstyle{definition}

\theoremstyle{plain}

\usepackage{tikz}
\usepackage[most]{tcolorbox}
\usepackage{hyperref}

\definecolor{lightblue}{RGB}{156, 193, 235}   
\definecolor{lightorange}{RGB}{255, 158, 99}     
\definecolor{lightpurple}{RGB}{202, 187, 253}
\definecolor{darkorange}{RGB}{255, 109, 77}
\definecolor{darkblue}{RGB}{81, 112, 255}
\definecolor{darkpurple}{RGB}{140, 82, 255}

\definecolor{superlightblue}{RGB}{207, 230, 255}
\definecolor{superlightgray}{RGB}{215, 215, 215}
\definecolor{superlightgreen}{RGB}{164, 217, 150}

\definecolor{forestgreen}{RGB}{34,139,34}   
\definecolor{firebrick}{RGB}{178,34,34}     
\newcommand{\cmark}{\textcolor{forestgreen}{\ding{51}}}
\newcommand{\xmark}{\textcolor{firebrick}{\ding{55}}}

\tikzset{
  paper/.style={
    draw,
    rectangle,
    rounded corners=5pt,
    align=center,
    minimum width=1.6cm,
    minimum height=0.6cm,
    font=\scriptsize
  },
  control/.style={paper, fill=forestgreen!50},
  planning/.style={paper, fill=darkorange!50},
  arrow/.style={
    ->,
    >=Stealth,
    thick,
    font=\scriptsize     
  }
}

\DeclarePairedDelimiterX{\norm}[1]{\lVert}{\rVert}{#1}

\usepackage[para,online,flushleft]{threeparttable}
\usepackage{array,booktabs,makecell}
\usepackage{bm}
\let\labelindent\relax
\usepackage[inline]{enumitem}
\usepackage{booktabs}
\usepackage{relsize}
\usepackage{accents}
\usepackage{array}
\usepackage{tabularx}
\usepackage{multirow}

\let\oldnl\nl
\newcommand{\nonl}{\renewcommand{\nl}{\let\nl\oldnl}}
\newcommand{\method}{\textbf{LAN2CB}\xspace}

\title{\LARGE \bf
Compositional Coordination for Multi‑Robot Teams with Large Language Models
}

\author{Zhehui Huang$^{1}$, Guangyao Shi$^{1}$, Yuwei Wu$^{2}$, Vijay Kumar$^{2}$, Gaurav S. Sukhatme$^{1}$
\thanks{
$^{1}$ Zhehui Huang, Guangyao Shi, and Gaurav S. Sukhatme are with the Department of Computer Science, University of Southern California, Los Angeles, CA 90089, USA. Email: \texttt{\{zhehuihu, shig, gaurav\}@usc.edu}.
$^{2}$ Yuwei Wu and Vijay Kumar are with the GRASP Lab, University of Pennsylvania, Philadelphia, PA 19104, USA. Email: \texttt{\{yuweiwu, kumar\}@seas.upenn.edu}.}
}

\begin{document}

\maketitle
\thispagestyle{empty}
\pagestyle{empty}

\begin{abstract}
 Multi-robot coordination has traditionally relied on a mission-specific and expert-driven pipeline, where natural language mission descriptions are manually translated by domain experts into mathematical formulation, algorithm design, and executable code. This conventional process is labor-intensive, inaccessible to non-experts, and inflexible to changes in mission requirements. Here, we propose \method (Language to Collective Behavior), a novel framework that leverages large language models (LLMs) to streamline and generalize the multi-robot coordination pipeline. \method transforms natural language (NL) mission descriptions into executable Python code for multi-robot systems through two core modules: (1) Mission Analysis, which parses mission descriptions into behavior trees, and (2) Code Generation, which leverages the behavior tree and a structured knowledge base to generate robot control code. We further introduce a dataset of natural language mission descriptions to support development and benchmarking. Experiments in both simulation and real-world environments demonstrate that \method enables robust and flexible multi-robot coordination from natural language, significantly reducing manual engineering effort and supporting broad generalization across diverse mission types. \textbf{Website:} 
\url{https://sites.google.com/view/lan-cb}\looseness=-1

\end{abstract}

\section{Introduction}

\noindent \textbf{Problem:} Multi-robot coordination for complex collective behaviors, including coverage, formation, foraging, and exploration, has been extensively studied for decades~\cite{844100, Turpin-2014-7831, Parker2016, 6979244, 7354037,amigoni2017multi, 8255576}. 
However, most prior coordination methods target a specific and narrowly defined coordination task. This 
highlights the need for a general and unified framework that can address a wide range of coordination problems in a systematic way.
The conventional multi-robot development pipeline starts with a natural language (NL) problem description, enabling both expert and non-expert users to specify objectives and constraints. 
An expert translates this description into a mathematical formulation, typically as an optimization problem, which inevitably requires additional assumptions.
Following this, a domain expert designs algorithms to solve the formulated problem, which are subsequently implemented and validated on robotic hardware.
However, this pipeline has three limitations: It is labor-intensive, inaccessible to non-experts, and inflexible. The first two limitations are apparent due to the significant use of skilled experts, despite the repetitive nature of problem formulation, algorithm design, and code implementation. 
The third becomes apparent when even minor changes to the problem statement or new requirements emerge, as the entire pipeline must be manually redesigned.

\noindent \textbf{Claim:} We posit that recent advances in large language models (LLMs) can be leveraged to alleviate these three limitations. Specifically, the use of such models can significantly reduce human expert effort and enhance the flexibility of multi-robot coordination systems~\cite{li2024improvingtenorlabelingreevaluating, li2025largelanguagemodelsstruggle, kasneci2023chatgpt, wu2024hierarchical, jiang2024survey}.

\noindent \textbf{Background:} LLMs are trained on massive amounts of text data to understand and generate human-like language, allowing them to interpret natural language prompts, perform contextual reasoning, and produce coherent responses or actions in both visual and linguistic tasks\cite{zeng2023large, wei2022chain, gao2024physically, Li_2025_CVPR, li2025videohalluevaluatingmitigatingmultimodal, li2025semanticallyawarerewardsopenendedr1, li2024pedantscheapeffectiveinterpretable}.
In robotics, LLMs are increasingly being used to improve high-level decision-making, task planning, and human-robot interaction~\cite{10885890, obata2024lip, shah2022lmnav, 10.1609/aaai.v38i17.29858, ravichandran_spine, sinha2024real,  khan2025safetyawaretaskplanning, mandi2024roco, kannan2024smart}. 
Applications include interpreting natural language instructions, generating structured plans, assisting in code generation for robot control, and enabling dialogue-based coordination among multi-robot systems. The key advantages of LLMs in robotics include their flexibility, ability to adapt to new tasks without extensive retraining, and capacity to bridge the gap between human intent and machine execution, thereby significantly reducing the need for manual programming and domain-specific engineering.

\begin{figure}[!t]
    \centering
\includegraphics[width=0.48\textwidth]{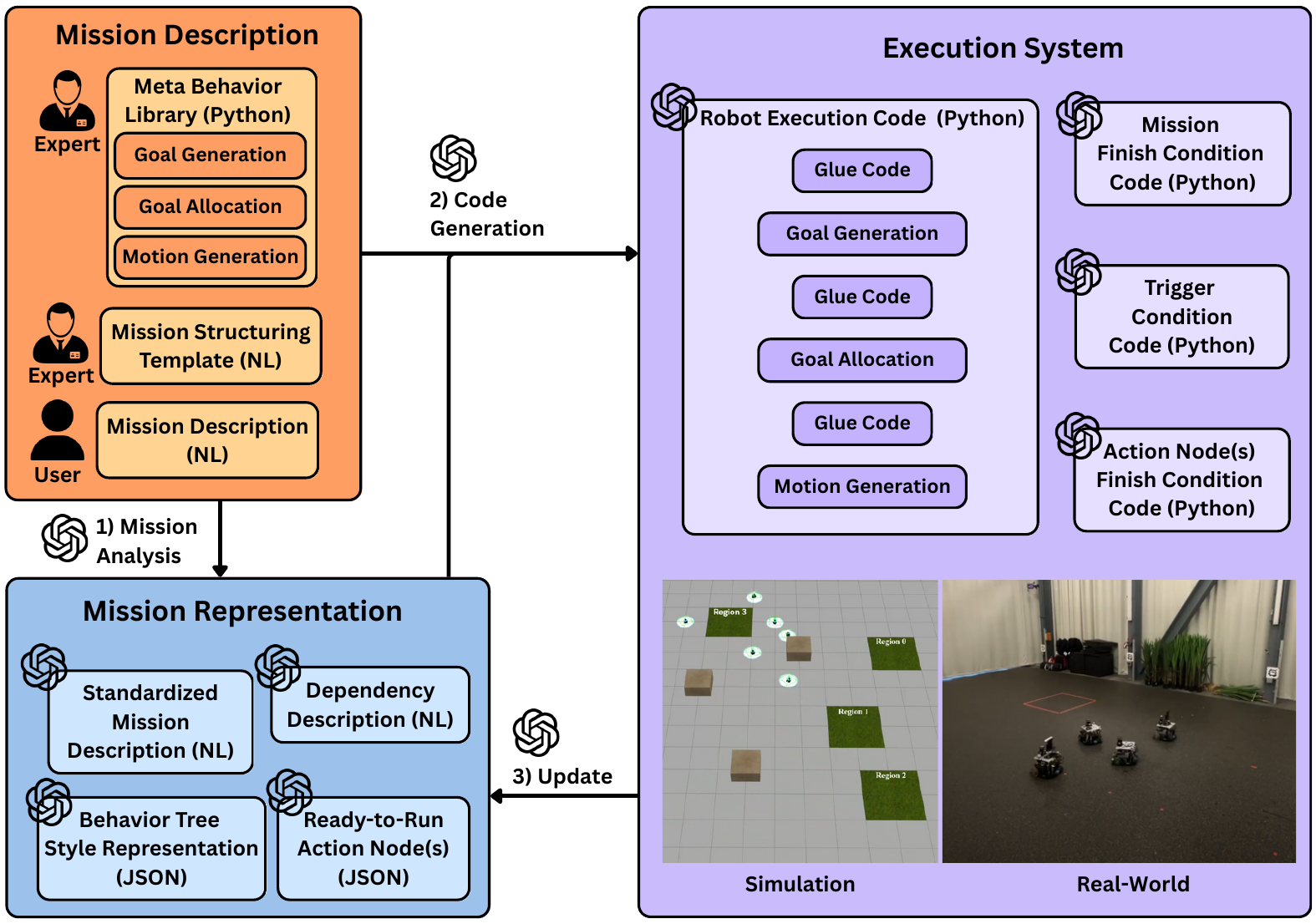}
    \vspace{-0.7cm}
    \caption{
    \method is an LLM-assisted multi-robot framework that converts mission descriptions into structured representations (e.g., behavior trees), assigns roles and priorities to each robot, and automatically generates executable code to accomplish complex missions.
    A mission change does not require human intervention since new code is generated automatically. NL means natural language. 
    \colorbox{lightorange}{\phantom{\rule{6pt}{2pt}}} user- or expert-provided content.
    \colorbox{lightblue}{\phantom{\rule{6pt}{2pt}}} LLM-generated mission representations.
    \colorbox{lightpurple}{\phantom{\rule{6pt}{2pt}}} LLM-generated code.
    }
    \label{fig:main_framework}
    \vspace{-0.81cm}
\end{figure}

\noindent \textbf{Contributions:} 1. We propose, implement, and thoroughly evaluate an LLM-based framework, Language to Collective Behavior (\method), which addresses the limitations of the traditional pipeline and turns natural language mission descriptions into executable plans and Python code for robot teams. \method consists of two main modules (Fig.~\ref{fig:main_framework}). The first module is \textsc{Mission Analysis}, which takes the mission specified in natural language as input, identifies the necessary tasks to be completed, and analyzes the dependency between tasks. The output of the first module, all ready-to-run nodes from a behavior tree, is given to the second module \textsc{Code Generation}, together with a pre-constructed knowledge base, to generate executable code for robots. 
2. We design a dataset of natural language mission specifications for multi-robot coordination. 
3. Our experiments, both in simulation and hardware, demonstrate that \method produces effective multi-robot coordination from natural language. They also show that \method is flexible and extendable, obviating the need to design from scratch whenever the mission changes and providing a standard format to incorporate existing knowledge into its knowledge base.

\section{Related Work}

\subsection{Multi-Robot Coordination}

Multi-robot systems can solve large tasks efficiently through cooperation, coordination, or collaboration~\cite{prorokresilience}.
Prior work has shown that these approaches enable robots to contribute effectively to overall team performance, either through global control strategies~\cite{680621} or by relying on local observations and decentralized decision-making~\cite{iocchi2003distributed, 4543197}.
As tasks become more complex, task-level coordination has received growing attention, where high-level goals are decomposed into interdependent sub-tasks that are allocated and executed collaboratively by the team~\cite{9410352, 9663414, 9013090}.
For example, Messing et al.~\cite{doi:10.1177/02783649211052066} introduced a four-layer structure covering task planning, allocation, scheduling, and motion planning to simplify decision-making. 
More recent work applied learning-based methods to multi-robot coordination by leveraging neural architectures such as graph attention networks for scheduling~\cite{9116987} and heterogeneous policy networks for communication and collaboration in diverse robot teams~\cite{10606072}.
However, most existing approaches remain limited to specific tasks and lack the generalization to handle more dynamic objectives, understand natural language, or perform higher-level reasoning abilities.

\subsection{Integrating LLMs in Robotics}

\subsubsection{Single-Robot Integration}

Recent work has explored integrating LLMs with single robots to enable more intuitive human-robot interaction and high-level task planning. 
These approaches leverage the language understanding and reasoning capabilities of LLMs to interpret natural language commands and translate them into executable actions.
Because natural language is inherently tied to contexts and semantics, LLMs are particularly effective in understanding tasks such as navigation, exploration, and instruction following. 
In these scenarios, LLMs can help robots ground language inputs into spatial or semantic representations that guide behavior in complex, partially known environments~\cite{10.1609/aaai.v38i17.29858, ravichandran_spine, shah2022lmnav}.
In addition, real-time resilience~\cite{10885890} and anomaly detection~\cite{sinha2024real} have been integrated into LLM-based robotic pipelines to improve robustness in dynamic and unforeseen situations.

\subsubsection{Team-Level Integration}
Extending LLMs to multi-robot systems introduces challenges in coordination, communication, and scalability, with research still in the early stages exploring their use for flexible, robust multi-robot coordination~\cite{li2025large}.
Some works use LLMs to translate natural language into formal representations~\cite{chen2024autotamp, wei2025hierarchical, zhang2025lammapgeneral, liu23lang2ltl}, but they are limited to simple missions and cannot handle multi-robot teams or decomposable long-horizon tasks.
Others propose prompt-based frameworks for multi-robot coordination using LLMs~\cite{mandi2024roco, kannan2024smart}. For instance, Zhao et al. introduced RoCO~\cite{mandi2024roco}, where robots coordinate via interactive dialogues. However, such frameworks face limitations with large robot teams (e.g., $\geq 10$) and complex language-based missions.
Shyam et al. proposed SMART-LLM~\cite{kannan2024smart}, an LLM-based task planner for multi-robot teams with task decomposition and allocation, but it lacks domain knowledge, limits generalization beyond in-context examples, and cannot handle motion-level language commands.
Our work is closely related to GenSwarm~\cite{ji2025genswarm}, which translates natural language task descriptions into executable Python code. While both aim to enable language-driven multi-robot coordination, our framework differs in two key ways. First, our modular design supports a broader range of coordination problems, including multi-team tasks with interdependencies and trigger events, unlike GenSwarm’s focus on simple, single-team scenarios. Second, we incorporate a structured knowledge base to guide the LLM, enhancing extensibility, whereas GenSwarm relies solely on in-context learning, limiting generality.

\begin{figure*}[!t]
    \centering
        \vspace{0.15cm}
\includegraphics[width=0.82\textwidth]{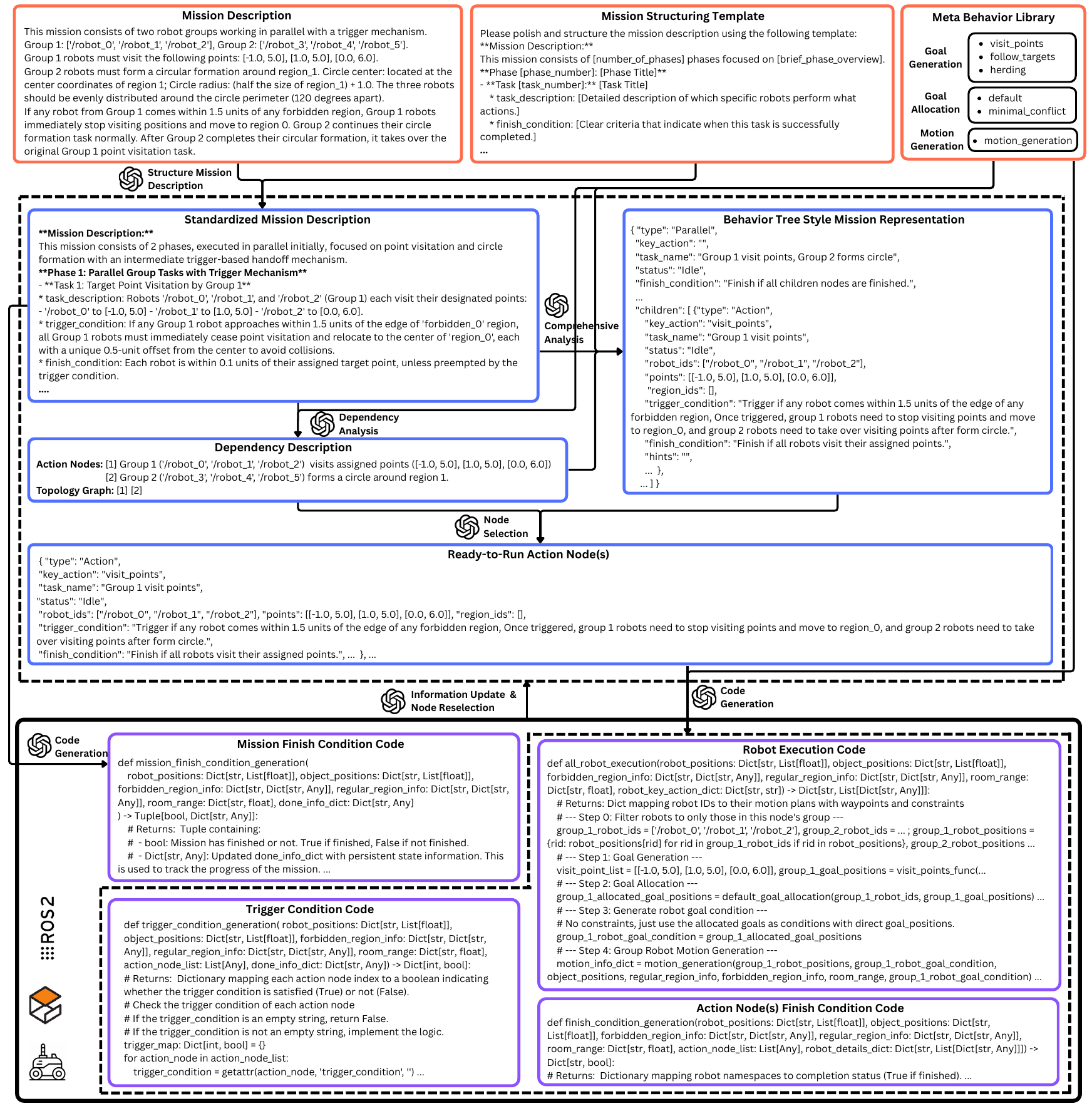}
    \vspace{-0.4cm}
    \caption{A comprehensive  demonstration of \textbf{LAN2CB}. {\setlength{\fboxrule}{1pt}
    \fcolorbox{darkorange}{white}{\phantom{\rule{5pt}{2pt}}} human-provided content.
    \fcolorbox{darkblue}{white}{\phantom{\rule{5pt}{2pt}}}  LLM-generated mission descriptions.
    \fcolorbox{darkpurple}{white}{\phantom{\rule{5pt}{2pt}}} LLM-generated code.}
    }
    \label{fig:workflow}
    \vspace{-0.62cm}
\end{figure*}

\section{Problem Statement}
\label{sec:problem}
Consider an environment with $N$ robots. 
Given natural language mission descriptions from users, our goal is to develop a system that can automatically control robots to accomplish these missions without further human intervention. 
We focus on two fundamental categories of missions: 1) goal-approaching missions, such as geometric formation or region visitation, and
2) herding missions, where robots must guide dynamic objects into specific regions or achieve a specific coverage percentage of an area.
Mission diversity is introduced by systematically varying key dimensions, resulting in a broad spectrum of possible tasks:
\begin{enumerate}
\item \textbf{Coordination Strategies:} Missions may require robots to coordinate, form a range of geometric patterns (e.g., lines, circles, characters), visit assigned regions, track dynamic targets, or herd dynamic objects.
\item \textbf{Motion Patterns:} Robots may be required to execute various motion trajectories, such as moving in straight lines, zigzagging, or spiral paths. 
\item \textbf{Constraints:} Additional requirements can be imposed, such as entering a region from a specific direction or adhering to prescribed durations for actions. 
\item \textbf{Trigger:} Tasks can include trigger conditions that allow actions to be terminated or modified in response to environmental events. For example, if a robot comes within a specified distance (e.g., 1 meter) of a forbidden region, it needs to switch to another task.
\item \textbf{Task Finish Condition:} Each atomic task has a clearly defined completion criterion, such as reaching a designated position within a user-specified tolerance or herding objects within a given range of the target region.  
\item \textbf{Mission Finish Condition:} Mission completion is defined more broadly, either by the successful completion of all constituent atomic tasks or by satisfying higher-level objectives (e.g., achieving at least 60\% coverage of a specified region).
\end{enumerate}

\vspace{-0.1cm}
\section{Method}
The \method workflow consists of the following stages, illustrated in Fig.~\ref{fig:main_framework}: 
\begin{enumerate}
\item \textbf{Mission Analysis}: Given natural language mission descriptions from users, the LLM autonomously decomposes missions into atomic task specifications and analyze inter-task dependencies. For each task, the LLM extracts robot assignments, relevant constraints, triggers, finish conditions, contextual hints, and other contextual requirements. Tasks that are immediately executable are then identified and forwarded to the next stage. 
\item \textbf{Code Generation}: Based on the mission description, the system first generates the mission finish condition, which is generated once for the entire mission cycle. For each selected task, considering its constraints, trigger conditions, finish conditions, and hints, the system separately generates code for: (a) robot execution, (b) trigger conditions, and (c) task finish conditions. 
\item \textbf{Execution}: The generated code is deployed to the robots for execution. During run-time, if any trigger condition or task finish condition is met, the workflow transitions to stage 4. If the overall mission finish condition is satisfied, it proceeds directly to stage 5. 
\item \textbf{Mission Progress Update}: Upon receiving execution results, the system uses the LLM to update the status of each task, then returns to stage 1 for further analysis and potential replanning if necessary.
\item \textbf{Mission Completion}: Once the mission finish condition is met, the mission is formally marked as complete.
\end{enumerate}

\subsection{Mission Analysis}
Mission analysis consists of two key phases: mission understanding and ready-to-run action node selection. 
However, user-provided natural language mission descriptions often omit explicit definitions of atomic tasks and their execution dependencies. As a result, directly prompting LLMs for general mission understanding can yield incomplete or ambiguous task specifications and may overlook critical execution dependencies. 
To address these challenges, the mission understanding phase begins by standardizing the mission description using an expert-provided template. We then perform dependency analysis to explicitly extract task relationships, which serve as the basis for a comprehensive mission analysis. 
Afterward, the LLM selects ready-to-run action nodes based on the results of the dependency and comprehensive mission analyses.

\textbf{Dependency Analysis}:
To initiate dependency analysis, we prompt the LLM to extract all atomic tasks from the mission description, initially disregarding trigger conditions and any tasks that may be activated post-trigger. The LLM then analyzes the dependency relationships among these atomic tasks. If triggers are activated during execution, we re-invoke the LLM to update the dependency analysis accordingly.

\textbf{Comprehensive Mission Analysis}:
Building on the results of dependency analysis, we combine these atomic tasks and their dependencies with the standardized mission description to prompt the LLM for comprehensive mission analysis. The mission is represented in a behavior tree structure, with the overall mission as the root and atomic tasks as action nodes. Non-leaf nodes of type \texttt{Parallel} or \texttt{Sequence} capture task execution topologies according to dependencies.
For each node, we instruct the LLM to generate a unique index, identify the node type, task name, status, constraints, trigger conditions, finish conditions, hints, and children. For action nodes, the LLM also specifies the action type as well as the associated robots, objects, regions, and points.
See Tab.~\ref{tab:behavior_tree_node} for detailed field descriptions of the \texttt{BehaviorTreeNode}.

\begin{table}
\centering
\vspace{0.2cm}
\caption{Field Descriptions for \texttt{BehaviorTreeNode}.}
\label{tab:behavior_tree_node}
\vspace{-0.3cm}
\resizebox{.44\textwidth}{!}{%
  \begin{tabular}{|>{\centering\arraybackslash}m{.15\textwidth}|>{\arraybackslash}m{.31\textwidth}|}
    \hline
    \textbf{Field (All Nodes)} & \textbf{Description} \\
    \hline
    \text{idx} & Unique index of the node. \\
    \hline
    \text{node\_type} & \text{Parallel} $|$ \text{Sequence} $|$ \text{Action}. \\
    \hline
    \text{task\_name} & \text{Concise task description of the node}. \\
    \hline
    \text{status} & \text{Idle} $|$ \text{Running} $|$ \text{Success} $|$ \text{Failure}. \\
    \hline
    \text{constraints} & Constraints such as max speed, time limit, regions to avoid, and etc. \\
    \hline
    \text{trigger\_condition} & Condition(s) that trigger node termination and activate new nodes. \\
    \hline
    \text{finish\_condition} & Condition(s) for determining task completion. \\
    \hline
    \text{hints} & Hints or suggestions. \\
    \hline
    \text{children} & List of child \text{BehaviorTreeNode} objects (empty for Action nodes). \\
    \hline
    \textbf{Field (Action Nodes)} & \textbf{Description} \\
    \hline
    \text{action\_type} & \text{visit\_points} $|$ \text{follow\_targets} $|$ \text{herd}. \\
    \hline
    \text{robot\_ids} & List of robot IDs this action node will control. \\
    \hline
    \text{object\_ids} & List of object IDs related to the action node. \\
    \hline
    \text{region\_ids} & List of region IDs related to the action node. \\
    \hline
    \text{points} & List of points related to the action node. \\
    \hline
  \end{tabular}%
}
\vspace{-0.7cm}
\end{table}

\textbf{Ready-to-Run Action Node(s) Selection}:
After constructing the complete behavior tree and integrating the dependency analysis results, we ask the LLM to identify action nodes that are immediately executable. Specifically, these are nodes for which all preconditions are met. The ready-to-run action nodes are then selected and forwarded for downstream processing, code generation. This approach ensures that robot actions are initiated only when all prerequisite tasks and constraints have been resolved, thereby supporting safe and efficient mission execution.
After selecting action nodes that are ready-to-run, we instruct the LLM to update the status of all nodes in the behavior tree, marking the selected nodes and their parent nodes as \textit{Running}.

\subsection{Code Generation}
After mission analysis and the identification of ready-to-run action nodes, we prompt the LLM to generate Python code to control all robots in the environment. 
The code generation process consists of four key components, executed synchronously with the robots' control loop (Fig.~\ref{fig:main_framework} and ~\ref{fig:workflow}):
1) Robot execution code generation;
2) Trigger condition code generation;
3) Action node finish condition code generation; 
4) Mission finish condition code generation.

\textbf{Robot Execution Code Generation}:
Recent works have shown that LLMs can effectively assist code generation~\cite{singh2023progprompt, ahn2022can, liang2023code, meng2025audere}. However, to improve success rates and mitigate hallucinations, it is more effective to provide core functions in advance and expose them as APIs, rather than asking the LLM to generate all code from scratch~\cite{huang2024can}. 
To this end, we construct a meta behavior library for robot execution.
To ensure flexibility and the capability to process multiple action nodes simultaneously, the meta-behavior library comprises three sub-libraries:
\begin{itemize}
\item \textit{Goal Generation}: Provides primitives for three fundamental behaviors: (a) visiting points, (b) following targets, and (c) herding.
\item \textit{Goal Allocation}: Offers two strategies for mapping generated goals to robots: (a) a default mapping that assigns goal positions based on robot IDs, and (b) a minimal conflict strategy that assigns goals to minimize trajectory intersections, assuming straight-line movement from start to goal.
\item \textit{Motion Generation}: Generates feasible paths from start to goal, considering environmental constraints. For each waypoint, the function not only outputs the position but also constraints such as maximum speed for approach.
\end{itemize}
This modular design allows the LLM to compose robust execution code for all selected action nodes, promoting reusability, flexibility, and scalability in multi-robot systems. 

\textbf{Trigger Condition Code Generation}:
For each action node, LLMs are prompted to implement the corresponding trigger condition. 
If no trigger is specified, the function always returns \texttt{False}. 
Otherwise, the LLM encodes the trigger logic using available environmental information, including robot and object states and region properties. 

\textbf{Action Node Finish Condition Code Generation}: 
For each selected action node, the LLM generates code to evaluate whether the node’s specified finish condition has been met. For example, if the task is to visit three points, the action node is considered complete when all three points have been visited. 

\textbf{Mission Finish Condition Code Generation:} 
Mission finish condition generation is similar in nature to composite task finish condition generation. However, the LLM only needs to generate the mission finish condition code once for the entire mission.

\subsection{Execution}
Once all code scripts are generated, the system initiates or resumes execution.
During execution, if a trigger condition is satisfied, we prompt the LLM to update the dependency analysis by removing the current task that caused the trigger and adding any new tasks associated with the trigger. The updated dependency analysis, together with the mission description, is then used to prompt the LLM to generate a new behavior tree. The LLM subsequently selects new ready-to-run action nodes and generates code for them. 
If the finish condition of any action node is satisfied, the LLM updates the behavior tree. This may also include updating parent nodes if their completion depends solely on the completion of their children. The LLM then selects new ready-to-run action nodes and generates the corresponding code. 
If the finish condition of any composite node is satisfied, the LLM updates the behavior tree by marking all of its child nodes that have not started or are still running as \texttt{Failure}, and marking the composite node as \texttt{Success}. 
Execution proceeds iteratively until the mission finish condition is satisfied.\looseness=-1
\begin{figure*}[!t]
  \centering
  \vspace{0.15cm}
\includegraphics[width=0.75\textwidth]{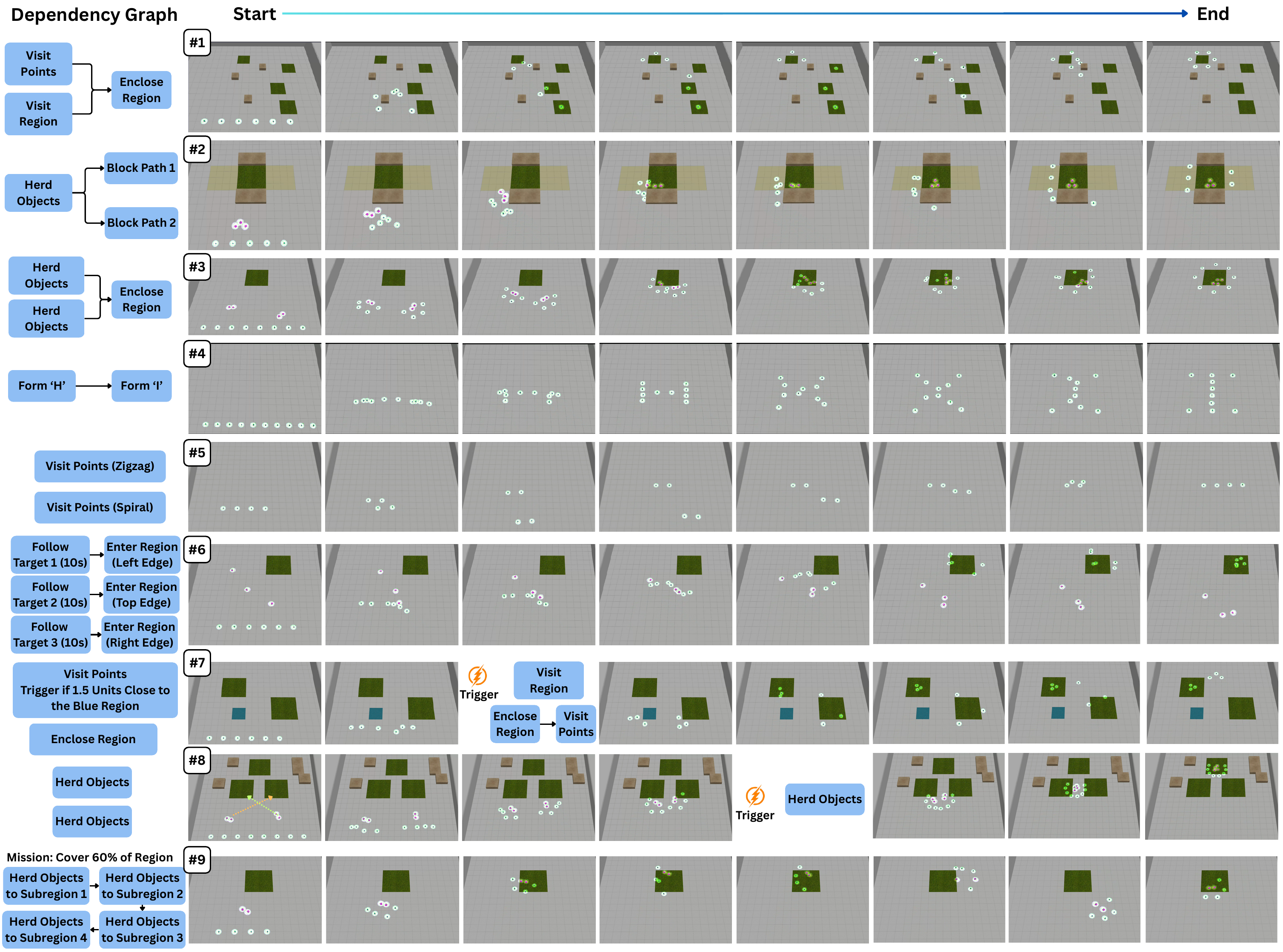}
  \vspace{-0.4cm}
  \caption{Quantitative results for nine multi-robot missions.  
  Please check the \href{https://sites.google.com/view/lan-cb/home}{website} for more details.
  }
  \vspace{-0.7cm}
  \label{fig:result_1}
\end{figure*}

\section{Experiments}
The experimental study is designed to rigorously evaluate \method under a variety of multi-step, long-horizon missions. We focus on three questions: 
1) How effectively does \method perform across diverse missions? 
2) How does the use of a standardized template for mission descriptions impact the performance of \method compared to raw natural language input?  
3) Can \method be reliably deployed on real-world robotic platforms? 

\subsection{Performance Across Diverse Missions}
\label{subsec:rq1}

To the best of our knowledge, no standard dataset exists for evaluating multi-step, long-horizon multi-robot mission execution under natural language mission descriptions. Therefore, we curate a comprehensive mission suite comprising three groups (nine missions total), each designed to test the distinct capabilities of our system.

\textbf{Mission Dataset}: 
We organize the missions by the complexity of mission analysis and required code generation: 
\begin{itemize}[leftmargin=*]
    \item \textit{Category A: Basic Missions}.
    These missions comprise multiple atomic tasks executed in a straightforward order, with no trigger conditions. Mission analysis and code generation are direct: all necessary information can be extracted from the natural language descriptions, enabling the LLM to generate input parameters for behavior library functions without additional reasoning. \emph{Examples: Assigning robots to visit predefined points.} (See Missions \#1, \#2, and \#3 in Fig.~\ref{fig:result_1}.)
    
    \item \textit{Category B: Basic Missions w/ Advanced Code Generation}.
    While maintaining a simple overall structure, these missions require advanced code generation. Certain behaviors are not directly supported by the behavior libraries, necessitating that the LLM synthesizes additional logic or routines. \emph{Example: Generating a spiral trajectory for robots when only straight-line motion is natively supported.} (See Missions \#4, \#5, and \#6 in Fig.~\ref{fig:result_1}.)
    
    \item \textit{Category C: Advanced Missions}.
    These missions demand advanced analysis and complex code generation. They may involve intricate dependencies, trigger conditions, or non-trivial mission finish conditions, such as region coverage requirements, dynamic triggers, or adaptive behaviors. (See Missions \#7, \#8, and \#9 in Fig.~\ref{fig:result_1}.)
\end{itemize}

\textbf{Evaluation Metrics}: 
We quantitatively evaluate missions using the metrics: 
(1) \textit{Avg. Success Rate}: The proportion of missions completed successfully as intended, without human intervention.
(2) \textit{Avg. Token Number and Cost}: The total number of LLM tokens used per mission and related cost.
(3) \textit{Avg. Tokens per Error}: The average number of tokens consumed per error given the output tokens of the LLM. 
\begin{table}[!t]
\vspace{0.2cm}
  \setlength{\tabcolsep}{2.8pt}    
  \renewcommand{\arraystretch}{1.2}
  \small
  \caption{Quantitative results.}
  \label{tab:results_metrics_vertical}
  \vspace{-3mm}
  \centering
  \resizebox{0.92\columnwidth}{!}{%
  \begin{tabular}{|l|c|c|c|c|c|c|}
    \hline
    \multirow{2}{*}{\makecell[l]{\textbf{Mission}\\\textbf{Category}}}
      & \multirow{2}{*}{\makecell[c]{\textbf{W/}\\\textbf{Template}}}
      & \multirow{2}{*}{\makecell[c]{\textbf{Avg.}\\\textbf{Success}\\\textbf{Rate}}}
      & \multicolumn{2}{c|}{\makecell[c]{\textbf{Avg. Tokens}\\\textbf{($10^3$)}}}
      & \multirow{2}{*}{\makecell[c]{\textbf{Avg.}\\\textbf{Cost (\$)}}}
      & \multirow{2}{*}{\makecell[c]{\textbf{Avg. Tokens}\\\textbf{/ Err ($10^3$)}}} \\
    \cline{4-5}
    & & & \textbf{Input} & \textbf{Output} & & \\
    \hline
    \multirow{2}{*}{Category A}
      & \xmark  &   0.87     &   48.70    &  12.33      &  0.20    &    92.48  \\ \cline{2-7}
      & \cmark &   1.00     &   52.83    &    14.53    &  0.22    & -     \\ \hline
    \multirow{2}{*}{Category B}
      & \xmark  &   0.73     &   40.53    &   9.97     &   0.16   &   37.39   \\ \cline{2-7}
      & \cmark &   0.93     &   35.25    &   8.54     &  0.14    &  128.10    \\ \hline
    \multirow{2}{*}{Category C}
      & \xmark  &    0.73    &  44.24     &   10.06     &   0.17   &  37.73    \\ \cline{2-7}
      & \cmark &   0.87     &  58.12     &    14.86    &  0.24    &  111.45    \\ \hline
    \multirow{2}{*}{\textbf{Overall}}
      & \xmark  &   0.78     &  44.49     &   10.79     &   0.18   &   55.87   \\ \cline{2-7}
      & \cmark &   0.93    &   48.73    &   12.64    &   0.20   &   119.78   \\ \hline
      \multicolumn{4}{l}{*Avg.\ Tokens / Error is based on output tokens.} \\
  \end{tabular}
  }
  \vspace{-0.76cm}
\end{table}

\textbf{Results}: 
We evaluate all missions using GPT-4.1, conducting each mission five times. Detailed results are provided in Tab.~\ref{tab:results_metrics_vertical}. 
By standardizing raw mission descriptions from users with an expert-provided template, \method achieves a 100\% success rate on Category A (basic missions), a 93\% success rate on Category B (basic missions with advanced code generation) and 87\% success rate on Category C (advanced missions). In addition, the system demonstrates strong robustness, achieving $\sim 0.12$ million tokens per error across all missions.
These results demonstrate that \method reliably analyzes mission descriptions, including identifying task dependencies, extracting essential information, selecting ready-to-run nodes, and dynamically updating behavior trees. Notably, the high success rates in Category B and Category C highlight its ability to perform advanced code generation tasks. These tasks include synthesizing robot trajectories not natively supported by the behavior library, generating custom formations, creating complex execution conditions such as specifying approach directions or target-following duration, and generating both triggers and mission finish conditions. Overall, our results indicate that \method is both robust and adaptable for diverse multi-robot mission execution scenarios.

\begin{figure}[!t]
    \centering
     \vspace{0.05cm}
\includegraphics[width=0.42\textwidth]{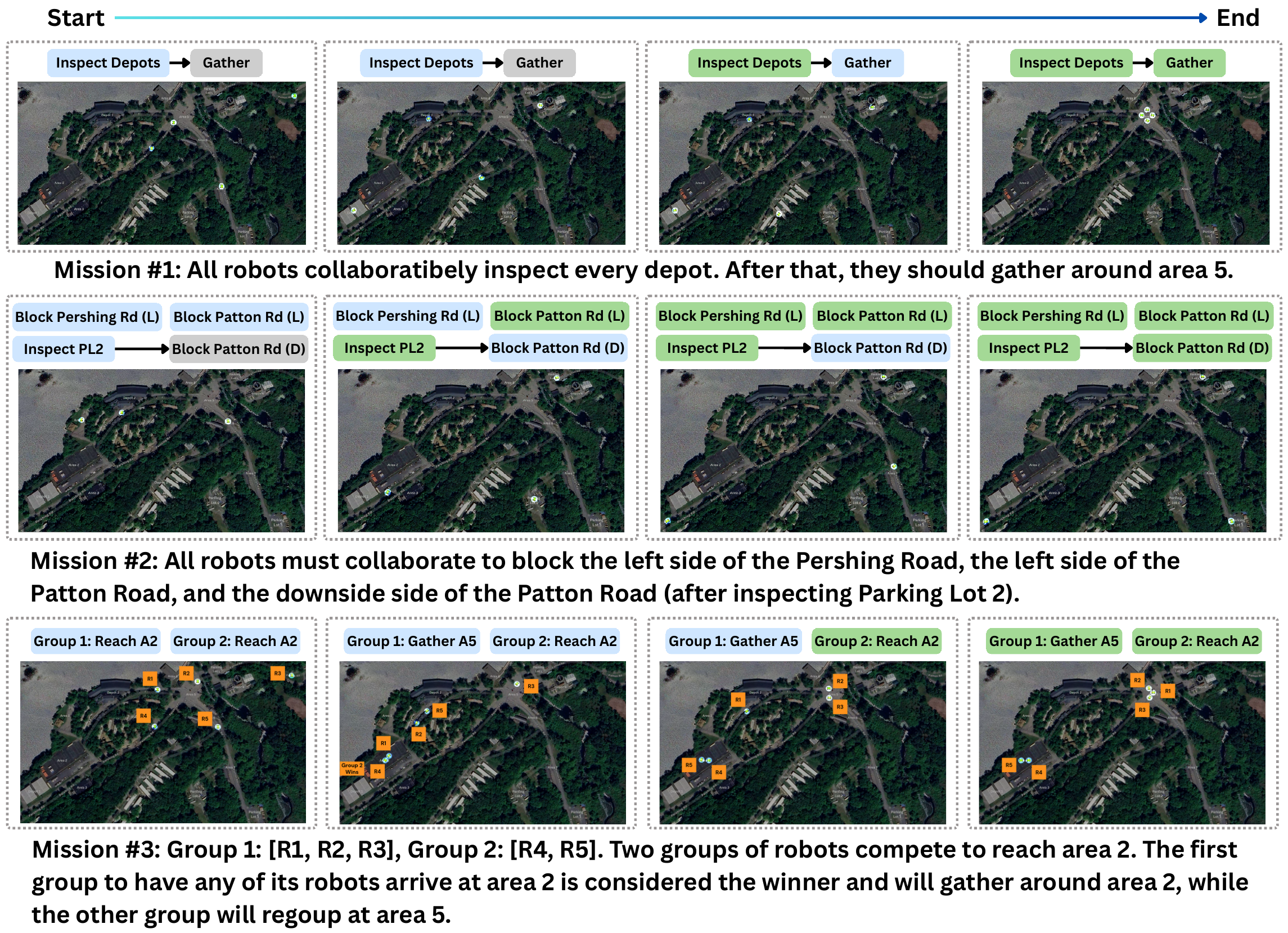}
    \vspace{-0.4cm}
    \caption{
    Realistic-setting experiments. 
    \colorbox{superlightgray}{\phantom{\rule{6pt}{2pt}}} idle,
    \colorbox{superlightblue}{\phantom{\rule{6pt}{2pt}}} running,
    \colorbox{superlightgreen}{\phantom{\rule{6pt}{2pt}}} finished.
    }
    \vspace{-0.86cm}
    \label{fig:dcist_v1}
\end{figure}

\subsection{Impact of standardized template}
To quantify the impact of using a template to standardize the mission descriptions on \method, we ablated the template-reformatting stage that aligns user-provided natural language mission descriptions to our predefined mission schema. As reported in Tab.~\ref{tab:results_metrics_vertical}, this omission results in an average drop of $\sim 15\%$ in success rates and causes the error rate to become $2.14\times$ more frequent across all mission categories.
These results highlight the critical role of structured prompting in maintaining robust multi-robot performance.\looseness=-1

\begin{figure}[!t]
    \centering
     \vspace{0.05cm}
\includegraphics[width=0.42\textwidth]{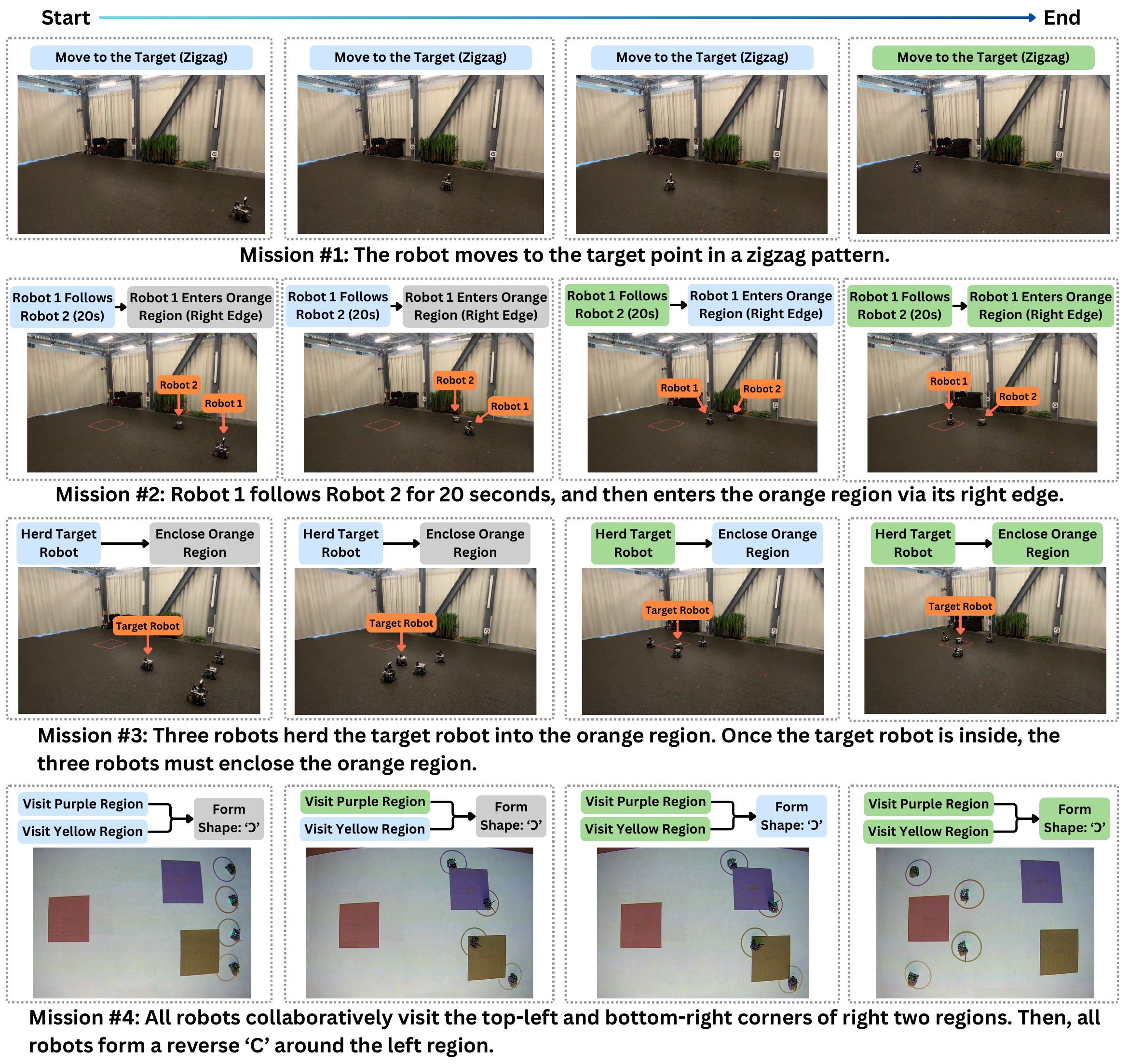}
    \vspace{-0.4cm}
    \caption{
    Real-world experiments. 
    \colorbox{superlightgray}{\phantom{\rule{6pt}{2pt}}} idle,
    \colorbox{superlightblue}{\phantom{\rule{6pt}{2pt}}} running,
    \colorbox{superlightgreen}{\phantom{\rule{6pt}{2pt}}} finished.
    }
    \vspace{-0.8cm}
    \label{fig:real_world_v1}
\end{figure}

\subsection{Realistic Simulated Scenarios}
We validated our framework in three simulation scenarios designed to  approximate real-world conditions (see Fig.~\ref{fig:dcist_v1}).

\subsection{Real World Demonstrations}
We validated our framework through four real-world demonstrations\footnote{We thank Jiazhen Liu for assistance with the physical deployment.} across two robot platforms (see Fig.~\ref{fig:real_world_v1}).

\section{Conclusions}

We presented an LLM-based framework, Language to Collective Behavior (\method), a pipeline that converts natural language into executable Python code for robot teams. Our experiments suggest that \method enables effective multi-robot coordination and remains flexible and extensible, even for multi-step and long-horizon missions. Additionally, we designed a dataset of natural language mission descriptions for multi-robot coordination.
For future work, we aim to improve mission analysis through retrieval-augmented generation (RAG)~\cite{10.5555/3495724.3496517}, enabling LLMs to leverage past mission data more effectively. Furthermore, we plan to expand the meta-behavior library to support a broader and more diverse range of capabilities.

\bibliography{MRS2025}

\end{document}